\documentclass[sigconf, nonacm]{acmart}
\usepackage{csquotes}
\usepackage{multirow}
\usepackage{graphicx}
\usepackage{placeins}

\AtBeginDocument{%
  }

\setcopyright{acmlicensed}
\copyrightyear{2018}
\acmYear{2018}
\acmDOI{XXXXXXX.XXXXXXX}

\acmConference[Conference acronym '26]{Make sure to enter the correct
  conference title from your rights confirmation email}{Apri, 2026}{Woodstock, NY}

\acmISBN{978-1-4503-XXXX-X/2018/06}

\begin{document}

\title{Resonance4D: Frequency-Domain Motion Supervision for Preset-Free Physical Parameter Learning in 4D Dynamic Physical Scene Simulation} 

\author{Changshe Zhang}
\email{zhangcs@stu.xidian.edu.cn}
\affiliation{%
  \institution{Xidian University}
  \country{China}
}

\author{Jie Feng}
\authornote{Corresponding author.}
\email{jiefeng0109@163.com}
\affiliation{%
  \institution{Xidian University}
  \country{China}
}

\author{Siyu Chen}
\email{chensy@ieee.org}
\affiliation{%
  \institution{Jimei University}
  \country{China}
}

\author{Guanbin Li}
\email{liguanbin@mail.sysu.edu.cn}
\affiliation{%
  \institution{Sun Yat-sen University}
  \country{China}
}

\author{Ronghua Shang}
\email{rhshang@mail.xidian.edu.cn}
\affiliation{%
  \institution{Xidian University}
  \country{China}
}

\author{Junpeng Zhang}
\email{junpengzhang@xidian.du.cn}
\affiliation{%
  \institution{Xidian University}
  \country{China}
}

\renewcommand{\shortauthors}{Zhang et al.}

\begin{abstract}

Physics-driven 4D dynamic simulation from static 3D scenes remains constrained by an overlooked contradiction: reliable motion supervision often relies on online video diffusion or optical-flow pipelines whose computational cost exceeds that of the simulator itself. Existing methods further simplify inverse physical modeling by optimizing only partial material parameters, limiting realism in scenes with complex materials and dynamics. We present Resonance4D, a physics-driven 4D dynamic simulation framework that couples 3D Gaussian Splatting with the Material Point Method through lightweight yet physically expressive supervision. Our key insight is that dynamic consistency can be enforced without dense temporal generation by jointly constraining motion in complementary domains. To this end, we introduce Dual-domain Motion Supervision (DMS), which combines spatial structural consistency for local deformation with frequency-domain spectral consistency for oscillatory and global dynamic patterns, substantially reducing training cost and memory overhead while preserving physically meaningful motion cues. To enable stable full-parameter physical recovery, we further combine zero-shot text-prompted segmentation with simulation-guided initialization to automatically decompose Gaussians into object-part-level regions and support joint optimization of full material parameters. Experiments on both synthetic and real scenes show that Resonance4D achieves strong physical fidelity and motion consistency while reducing peak GPU memory from over 35\,GB to around 20\,GB, enabling high-fidelity physics-driven 4D simulation on a single consumer-grade GPU.

\end{abstract}

\ccsdesc[500]{Computing methodologies~Reconstruction}
\ccsdesc[300]{Computing methodologies~Shape representations}
\ccsdesc[100]{Computing methodologies~3D imaging}

\keywords{Physics-based Simulation, Frequency-domain Motion Supervision, Dynamic 3D Gaussian Splatting}

\maketitle

\section{Introduction}

Generating physically consistent dynamic responses from static 3D reconstructions is a core problem in world modeling\cite{abou2024physically}, robotic interaction\cite{tseng2025gaussian}, and immersive content creation\cite{jiang2024robust}. 
With the success of 3D Gaussian Splatting (3DGS) in efficient rendering and novel-view synthesis \cite{kerbl20233d}, recent research has begun to integrate explicit 3D representations with differentiable physics simulation, aiming to recover material parameters from visual observations and generate interactive 4D dynamic scenes. 
Among these approaches, modeling frameworks based on the Material Point Method \cite{hu2018moving} have attracted particular attention due to their ability to uniformly handle large deformations, complex contacts, and diverse material behaviors. 
However, existing methods often depend on supervision strategies with high optimization overhead, while adopting simplified physical parameter settings that fall short of real-world material complexity.

On the one hand, existing physics-driven 4D simulation methods typically rely on online video diffusion models\cite{huang2025dreamphysics, zhang2024physdreamer, lin2025omniphysgs, liu2024physics3d} or optical flow networks\cite{liu2025unleashing} to enforce consistency between simulated results and reference motions. 
Although such strategies offer effective temporal supervision, they typically depend on additional large-scale auxiliary models, resulting in substantial GPU memory and computational overhead and limiting their use in long-horizon settings and on consumer-grade hardware. 
More importantly, purely spatial frame-wise matching is often insufficient to reliably capture long-range temporal dependencies in complex non-rigid motion, making optimization prone to temporal discontinuities and inadequate recovery of high-frequency oscillatory details. 
Our experiments further indicate that effective motion supervision does not inherently depend on large-scale online prior models. For physics-driven dynamic fitting, lightweight supervision that is more closely aligned with the underlying dynamics can still provide stable and competitive constraints, while substantially reducing optimization overhead.

On the other hand, physical parameter modeling in existing methods remains limited. 
Real-world non-rigid objects are rarely materially uniform: different parts often exhibit distinct stiffness, compressibility, plasticity, and mass distribution, and their dynamic behaviors are jointly governed by multiple coupled physical parameters. 
This characteristic has motivated recent physics-driven reconstruction methods to move beyond purely geometry-based motion fitting and incorporate explicit physical parameter optimization into dynamic modeling. 
However, as summarized in Table~\ref{tab:intro_support_phys}, most existing methods still optimize only a subset of physical parameters while fixing the others as empirical constants. 
Such partial modeling restricts the ability to capture spatially varying material properties and the coupled mechanisms underlying real-world deformation. 
Moreover, the choice of parameter-sharing granularity introduces another challenge. 
Object-level sharing is often too coarse to represent heterogeneous materials across different parts, whereas particle-level optimization introduces excessive degrees of freedom and becomes highly under-constrained under limited-view supervision. These limitations suggest that accurate physical modeling requires both more complete parameter recovery and a more flexible part-level parameterization.

\begin{figure}[t]
    \centering
    \includegraphics[width=\linewidth]{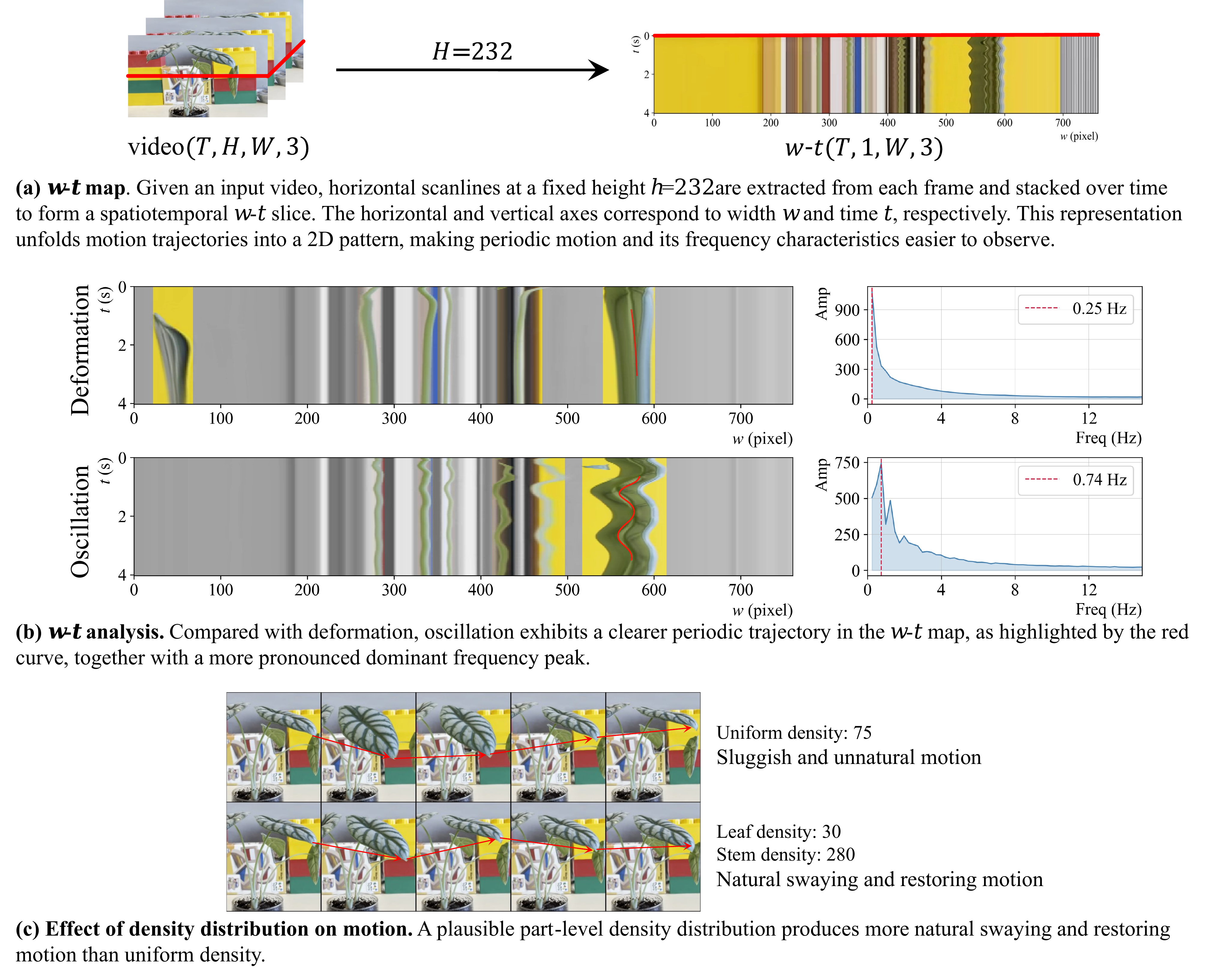}
    \caption{Motivation for dual-domain motion analysis and part-level physical modeling.
        (a) By extracting horizontal scanlines at a fixed height and stacking them over time, an input video is converted into a $w-t$ spatiotemporal slice that explicitly unfolds motion trajectories into a 2D structural pattern.
        (b) In this representation, oscillation exhibits a clearer periodic trajectory and a more pronounced dominant frequency peak than deformation.
        (c) Different density settings lead to markedly different motion behaviors: compared with uniform density, a more plausible part-level density distribution produces more natural swaying and restoring motion.}
    \Description{The figure contains three panels. Panel (a) shows how a $w$-$t$ spatiotemporal slice is formed by extracting a horizontal scanline at a fixed height from each video frame and stacking these scanlines over time, with width on the horizontal axis and time on the vertical axis. Panel (b) shows two examples of $w$-$t$ maps and their corresponding frequency spectra: a deformation example with smoother, less regular trajectories and a lower dominant frequency peak, and an oscillation example with a clearer periodic trajectory marked by a red curve and a higher dominant peak. Panel (c) shows two motion sequences generated with different density settings. The top sequence uses uniform density. The bottom sequence uses different densities for the leaf and stem, producing a larger swaying amplitude and a clearer restoring trend over time.}
    \label{fig:result2}
\end{figure}

\begin{table}[t]
    \centering
    \caption{Comparison of learnable physical parameters across different methods. A checkmark indicates that the corresponding parameter is explicitly optimized, while a blank entry indicates that it is not updated.}
    \label{tab:intro_support_phys}
    \resizebox{\linewidth}{!}{
        \begin{tabular}{lcccccc}
            \toprule
            \textbf{Method} & \multicolumn{5}{c}{\textbf{Constitutive Parameters}} & \textbf{Physical Property} \\
            \cmidrule(lr){2-6} \cmidrule(lr){7-7}
            {}              &Young's        &Poisson's      &Yield          &Plastic        &Friction       &\multirow{2}{*}{Density}   \\ 
            {}              &Mod.           &Ratio          &Stress         &Visc.          &Angle          &{}                         \\
            \midrule
            PhysDreamer\cite{zhang2024physdreamer}  &$\checkmark$   &{}             &{}             &{}             &{}             &{}                         \\
            DreamPhysics\cite{huang2025dreamphysics}&$\checkmark$   &{}             &{}             &{}             &{}             &{}                         \\
            Physics3D\cite{liu2024physics3d}        &$\checkmark$   &$\checkmark$   &$\checkmark$   &$\checkmark$   &$\checkmark$   &{}                         \\
            PhysFlow\cite{liu2025unleashing}        &$\checkmark$   &$\checkmark$   &$\checkmark$   &$\checkmark$   &$\checkmark$   &{}                         \\
            Ours                                    &$\checkmark$   &$\checkmark$   &$\checkmark$   &$\checkmark$   &$\checkmark$   &$\checkmark$               \\
            \bottomrule
        \end{tabular}
    }
\end{table}

To address the above issues, we propose Resonance4D, a physics-driven 4D dynamic simulation framework built upon 3DGS and MPM. 
The key idea is to eliminate the reliance on large online models through lightweight yet complementary dual-domain supervision, while improving the completeness and stability of material inversion via part-level full-parameter optimization. 
Specifically, we introduce Dual-domain Motion Supervision (DMS): in the spatial domain, structural consistency constraints are used to model slowly varying deformations; in the spatiotemporal frequency domain, spectral consistency constraints are used to capture local oscillations and global dynamic patterns. 
This design eliminates the need to invoke online diffusion models or optical flow networks during training, thereby significantly reducing optimization memory overhead while providing a complementary signal beyond frame-wise supervision for long-horizon dynamic fitting.
To enable parameter optimization, we further establish a part-level differentiable optimization, in which material properties and constitutive parameters are jointly incorporated into end-to-end learning. 
To improve the practicality of this process in real scenes, we introduce two key designs. 
First, we adopt part-level parameter sharing to balance expressive power and optimization stability. 
Second, we employ simulation-driven automatic initialization, which selects more reasonable starting points through rapid candidate simulations and metric-based screening, thereby reducing the sensitivity of full-parameter optimization to initialization. 
To reduce manual annotation cost, we also introduce zero-shot segmentation-based part extraction to automatically construct parameter-sharing units.

The main contributions of this paper are as follows:
\begin{itemize}

    \item Existing physics-driven 4D generation methods often rely on heavy online video priors for motion supervision, leading to high memory and optimization cost. In contrast, our study shows that effective dynamic supervision can be achieved with lightweight constraints in the spatial and spectral domains.
    \item Motivated by this observation, we propose Resonance4D, a unified physics-driven framework for 4D dynamic scene simulation with differentiable physical parameter learning from video observations. At its core, Dual-domain Motion Supervision (DMS) combines spatial structural consistency and spatiotemporal spectral consistency, removing the need for online diffusion or optical-flow models while reducing GPU memory overhead.
    \item For complex real scenes, we further develop a part-level full-parameter differentiable optimization strategy. By jointly learning material properties and constitutive parameters, together with simulation-driven initialization and zero-shot segmentation-based part extraction, the framework improves the stability and accuracy of physical parameter inversion in heterogeneous scenes.

\end{itemize}
Experiments on synthetic and real scenes demonstrate competitive performance in physical parameter recovery, motion consistency, and visual quality, while substantially lowering the hardware barrier.

\section{Related Work}

\subsection{3D/4D Scene Representation and Dynamic Reconstruction.}
Recovering renderable 3D scene representations from multi-view images or videos is a fundamental problem in visual computing.
Early methods such as Neural Radiance Fields (NeRF) \cite{mildenhall2021nerf} model scene geometry and appearance with continuous implicit fields, while 3D Gaussian Splatting (3DGS)\cite{kerbl20233d} introduces an explicit Gaussian-based representation that offers favorable training efficiency, rendering speed, and editability. 
These advances provide the foundation for dynamic scene modeling.
Building on static 3D representations, a large body of work extends NeRF and 3DGS to dynamic or 4D settings\cite{pumarola2021d, park2021nerfies, park2021hypernerf, luiten2024dynamic, wu20244d}. 
Methods such as D-NeRF\cite{pumarola2021d}, Nerfies\cite{park2021nerfies}, and HyperNeRF\cite{park2021hypernerf} model non-rigid scene dynamics through canonical spaces and learned deformation fields. 
More recent Gaussian-based approaches, including Dynamic 3D Gaussians \cite{luiten2024dynamic} and 4D Gaussian Splatting\cite{wu20244d}, represent scene evolution using temporally varying Gaussian primitives or deformation trajectories, achieving efficient and high-quality dynamic reconstruction and novel-view synthesis. 
However, these methods are primarily designed to reproduce observed time-varying geometry, appearance, and motion, rather than to infer the physical mechanisms or material parameters underlying the dynamics.

\subsection{Video Generation.}
Recent video generation has been dominated by diffusion-based and flow-matching models trained on large-scale video corpora. 
Representative methods such as Stable Video Diffusion\cite{blattmann2023stable} and VideoCrafter2\cite{chen2024videocrafter2} extend latent diffusion models to video generation through temporal modeling and staged training, while more recent systems such as CogVideoX\cite{yang2024cogvideox} and Sora\cite{liu2024sora} further scale this line with diffusion-transformer architectures, leading to substantial improvements in visual fidelity, motion realism, and temporal coherence. 
In addition, recent flow-matching approaches, such as Pyramidal Flow Matching\cite{jin2024pyramidal}, improve generation efficiency by modeling video synthesis with unified spatial-temporal pyramids. 
In parallel, recent autoregressive video models, such as FAR\cite{gu2025long} and LongLive\cite{yang2025longlive}, revisit causal next-frame generation and show strong potential for long-context or real-time video synthesis. 
Despite their strong generative capability, these methods model dynamics implicitly in latent spaces or neural generators, rather than through explicit physical states and interpretable material parameters. 
Nevertheless, the rich motion patterns learned by these models provide useful motion priors for physics-driven dynamic reconstruction and simulation.

\subsection{Physics-driven 3DGS.}
To improve physical consistency beyond purely appearance-driven dynamic modeling, recent work has begun to couple explicit scene representations with differentiable physics simulation. 
PhysGaussian is an early representative in this direction, integrating 3DGS with continuum mechanics and MPM simulation so that Gaussian primitives participate in both rendering and physical evolution\cite{xie2024physgaussian}. 
Subsequent works further explore the use of learned video priors for physical parameter estimation and dynamic simulation. 
PhysDreamer leverages motion priors from video simulation models to animate static 3D objects under interaction\cite{zhang2024physdreamer}. 
DreamPhysics introduces video diffusion priors into physical attribute learning for 3D Gaussians and combines them with MPM-based simulation for 4D dynamic synthesis\cite{huang2025dreamphysics}. 
PhysFlow further incorporates video diffusion, multimodal foundation models, and flow-based constraints to strengthen dynamic supervision and parameter optimization\cite{liu2025unleashing}. 
Other methods, such as OmniPhysGS\cite{lin2025omniphysgs} and PhysSplat\cite{zhao2025physsplat}, continue this direction by improving material modeling, scalability, or simulation efficiency. 
Despite encouraging progress, existing physics-driven 3DGS methods differ substantially in the form of supervision they require, the scope of physical parameters they optimize, and the degree to which they rely on external initialization, region decomposition, or auxiliary modules. 
These differences are particularly important for real-scene parameter inversion, where optimization stability, supervision cost, and controllability are central concerns.
\section{Method}

\begin{figure*}[t]
    \centering
    \includegraphics[width=1.0\textwidth]{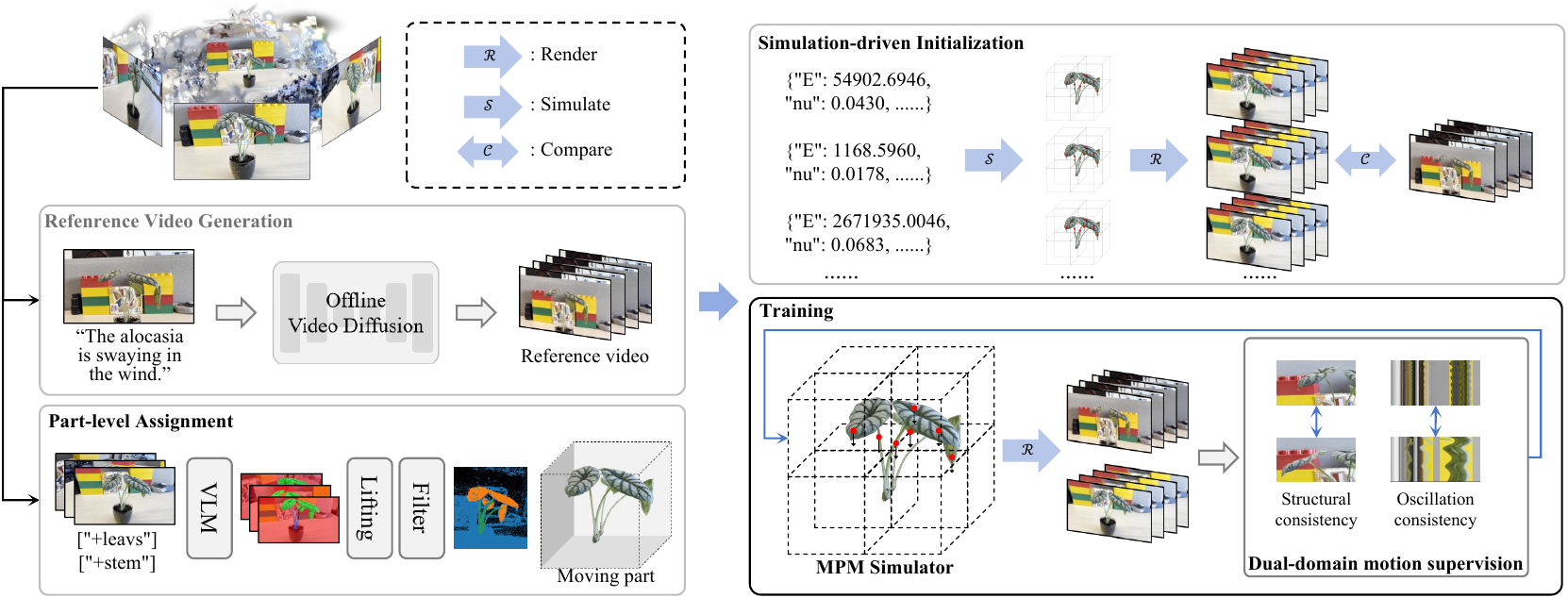}
    \caption{
    Resonance4D is a unified framework for physics-driven dynamic simulation from static reconstruction. 
    Its key insight is that effective motion supervision does not require heavy online video priors; lightweight dual-domain spatial-spectral constraints are sufficient to guide physically meaningful dynamics. 
    Guided by this insight, the framework first constructs a reference motion video, then identifies movable regions through automatic part-level assignment, next obtains a reliable starting point via simulation-driven initialization over the feasible physical parameter space, and finally performs part-level physical parameter optimization under Dual-domain Motion Supervision.
    }
    \Description{The figure presents a block diagram of the Resonance4D pipeline. A static reconstructed scene is shown on the left as input. A legend at the top indicates icons for rendering, simulation, and comparison. The middle-left area contains two modules: Reference Video Generation, which converts a motion text prompt into a reference video through an offline video diffusion model, and Part-level Assignment, which uses text prompts, a vision-language model, lifting, and filtering to obtain moving parts. The upper-right module, Simulation-driven Initialization, evaluates multiple candidate physical parameter sets by simulation, rendering, and comparison with the reference video. The lower-right module, Training, runs an MPM simulator on the object, renders the simulated frames, and applies dual-domain motion supervision based on structural consistency and oscillation consistency. Arrows connect the modules to indicate the order of processing and the optimization loop.}
    \label{fig:method}
\end{figure*}

Given a static 3D Gaussian scene, our goal is to recover physically meaningful material parameters and generate physically consistent 4D dynamics through explicit simulation. To this end, we propose a unified framework, illustrated in Figure~\ref{fig:method}, that tightly couples 3D Gaussian Splatting (3DGS) with a differentiable Material Point Method (MPM) simulator.

Our framework consists of three key components: dual-domain motion supervision for effective motion guidance, simulation-driven initialization for stabilizing optimization, and part-level joint physical parameter optimization, where moving-part extraction is used to define parameter-sharing regions. 
Specifically, Section~3.1 presents Dual-domain Motion Supervision (DMS), which provides lightweight yet effective constraints in the spatial and spectral domains. Section~3.2 introduces a simulation-driven initialization strategy that improves optimization stability by selecting a more reliable starting point. Section~3.3 describes the part-level joint physical parameter optimization framework, including moving-part extraction for defining parameter-sharing regions.

\subsection{Dual-domain Motion Supervision (DMS)}
A major challenge in physics-driven 4D simulation is how to provide effective motion supervision without relying on heavy online teachers such as diffusion models or optical-flow networks during optimization. 
Purely spatial supervision is often sufficient for coarse appearance alignment, but it may be less sensitive to local oscillatory motion patterns and fine-grained temporal inconsistencies. To address this issue, we complement spatial-domain supervision with a local spatiotemporal spectral loss.

let $\mathbf{V}^{*}=\{\mathbf{I}^{*}_t\}_{t=1}^{T}$ denote the reference video, and Let $\hat{\mathbf{V}}=\{\hat{\mathbf{I}}_t\}_{t=1}^{T}$ denote the rendered video from the current simulation parameters, We first impose a spatial reconstruction loss on individual frames:
\begin{equation}
    \mathcal{L}_{\mathrm{spatial}} = 1-\frac{1}{T}\sum_{t=1}^{T}\mathrm{SSIM}(\hat{\mathbf{I}}_t,\mathbf{I}^{*}_t) ,
    \label{eq:l_spatial}
\end{equation}
This frame-wise spatial term provides stable structural supervision, but is less sensitive to local oscillatory motion and fine-grained temporal inconsistencies.

Beyond frame-wise appearance similarity, we introduce spectral supervision on frame differences to capture local motion patterns. Specifically, we define the temporal difference sequences as
\begin{equation}
    \Delta \hat{\mathbf{I}}_t = \hat{\mathbf{I}}_{t+1} - \hat{\mathbf{I}}_{t}, \qquad
    \Delta \mathbf{I}^{*}_t = \mathbf{I}^{*}_{t+1} - \mathbf{I}^{*}_{t},
    \quad t = 1,\dots,T-1.
    \label{eq:frame_diff}
\end{equation}
Using frame differences suppresses static appearance cues and highlights motion-dependent temporal variations, making the subsequent spectral supervision more sensitive to local dynamic patterns.

Object motion is often spatially localized and typically occupies only part of the image, while large background regions are static or only weakly varying. Directly applying a global 3D FFT to the entire video volume would therefore mix motion-relevant signals with irrelevant spatial content, making local dynamic patterns less distinguishable in the frequency domain. To obtain more informative local spectra, we instead patchify the frame-difference volumes along the spatial dimensions before performing the Fourier transform. Specifically, after grayscale conversion, given $\Delta \hat{\mathbf{I}} \in \mathbb{R}^{(T-1)\times H \times W}$ and $\Delta \mathbf{I}^{*} \in \mathbb{R}^{(T-1)\times H \times W}$,
the spatial dimensions are partitioned into non-overlapping patches of size $H_p \times W_p$, yielding
\begin{equation}
    \hat{\mathbf{P}}, \mathbf{P}^{*} \in \mathbb{R}^{N \times (T-1) \times H_p \times W_p},
    \qquad
    N = \frac{H}{H_p}\frac{W}{W_p}.
\end{equation}
Each patch preserves the full temporal extent while restricting the spatial support to a local region. We then compute the 3D Fourier transform for each patch over the temporal and spatial axes
\begin{equation}
    \hat{\mathbf{F}}_n = \mathcal{F}(\hat{\mathbf{P}}_n), \qquad
    \mathbf{F}^{*}_n = \mathcal{F}(\mathbf{P}^{*}_n),
    \label{eq:patch_fft}
\end{equation}
where $\mathcal{F}(\cdot)$ denotes the discrete Fourier transform over $(t,h,w)$.

We decompose each spectrum into magnitude and phase-related components. The magnitude term aligns the spectral energy distribution of local dynamic patterns
\begin{equation}
    \mathcal{L}_{\mathrm{mag}} = \left\| \log(1+|\hat{\mathbf{F}}_n|) - \log(1+|\mathbf{F}^{*}_n|) \right\|_1 .
    \label{eq:l_mag}
\end{equation}

To further enforce local spectral alignment, we introduce a phase consistency term. We first normalize each complex spectrum as
\begin{equation}
    \tilde{\mathbf{F}}_n = \frac{\hat{\mathbf{F}}_n}{|\hat{\mathbf{F}}_n|+\epsilon}, 
    \qquad
    \tilde{\mathbf{F}}^{*}_n = \frac{\mathbf{F}^{*}_n}{|\mathbf{F}^{*}_n|+\epsilon},
    \label{eq:phase_norm}
\end{equation}
where $\epsilon$ is a small constant for numerical stability. The phase similarity map is then defined as
\begin{equation}
    \mathbf{S}_{\mathrm{phase}}^{(n)} = 
    \mathrm{Re}\!\left( \tilde{\mathbf{F}}_n \odot \overline{\tilde{\mathbf{F}}^{*}_n} \right),
    \label{eq:phase_similarity}
\end{equation}
with $\odot$ denoting element-wise multiplication. The corresponding phase loss is
\begin{equation}
    \mathcal{L}_{\mathrm{phase}} = 
    1 - \mathrm{mean}\!\left( \mathbf{S}_{\mathrm{phase}}^{(n)} \right),
    \label{eq:l_phase}
\end{equation}
where $\mathrm{mean}(\cdot)$ averages over all elements of the phase similarity tensor.

Finally, the spectral supervision term and the full dual-domain motion supervision are defined as
\begin{equation}
    \mathcal{L}_{\mathrm{spec}} = \mathcal{L}_{\mathrm{mag}} + \mathcal{L}_{\mathrm{phase}}, \qquad
    \mathcal{L}_{\mathrm{DMS}} = \mathcal{L}_{\mathrm{spatial}} + \mathcal{L}_{\mathrm{spec}} .
    \label{eq:l_dms}
\end{equation}

\subsection{Simulation-driven Initialization}
Joint optimization over multiple physical parameters is highly sensitive to initialization, since different parameter combinations may induce substantially different motion patterns while yielding similarly plausible frame-wise appearances. Instead of directly starting from a single hand-crafted guess, we first perform a coarse simulation-driven search to obtain a more reliable initialization.

Physical parameters often span several orders of magnitude. For example, Young's modulus may vary from $10^3$ to $10^7$, making linear sampling inefficient. We therefore perform Latin hypercube sampling (LHS) in log space. Let the valid range of the $j$-th parameter be $[l_j,u_j]$, and let $D$ be the number of optimized parameters. We sample $ \mathbf{u}_n \sim \mathrm{LHS}([0,1]^D, N) $ and map each sample to the physical parameter space as:
\begin{equation}
    s_{n,j} = \log_{10} l_j + \left(\log_{10} u_j - \log_{10} l_j\right) u_{n,j},
    \qquad
    \theta^{(n)}_j = 10^{s_{n,j}},
\label{eq:lhs_map}
\end{equation}
where $N$ is the number of sampled candidates.

For each candidate parameter vector $\boldsymbol{\theta}^{(n)}$, we run a short forward MPM simulation for $K$ frames and render the corresponding video $\hat{\mathbf{V}}_n$. We then select the parameter vector whose rendered motion is most similar to the reference video under a lightweight video similarity metric:
\begin{equation}
    \boldsymbol{\theta}^{*} = \arg\max_{\boldsymbol{\theta}^{(n)}} 
    \mathrm{MS\text{-}SSIM} (\hat{\mathbf{V}}_n,\mathbf{V}^{*}) .
    \label{eq:init_select}
\end{equation}
The selected parameter vector $\boldsymbol{\theta}^{*}$ is used to initialize the subsequent differentiable optimization.

\subsection{Part-level Joint Physical Parameter Optimization}
A key challenge in physical parameter inversion is to choose an appropriate parameter-sharing granularity. Object-level parameterization is often too coarse to capture spatially heterogeneous material properties, whereas particle-level optimization is typically under-constrained under limited visual supervision and thus prone to instability. To balance expressiveness and optimization stability, we perform joint physical parameter optimization at the part level. Concretely, this module consists of two stages. We first derive stable part-level assignments from multi-view observations without manual annotation, which define the parameter-sharing units for subsequent inversion. We then associate each part with a learnable physical parameter vector and optimize these parameters under motion supervision, starting from the simulation-driven initialization described above.

\noindent \textbf{Part-level Assignment.}
We obtain part-level assignments automatically from zero-shot text-prompted segmentation on the input multi-view images using DINOv3~\cite{dinov3} and SAM3~\cite{sam3}, producing 2D semantic masks for each view. We then lift these 2D segmentation results to the 3D Gaussian representation following LUDVIG~\cite{marrie2025ludvig}. Specifically, for each Gaussian, we aggregate its category responses across views and apply a softmax function to estimate the probability of belonging to each part category, assigning the most likely label accordingly. Finally, we apply a KNN-based post-processing step to suppress isolated noisy points and improve local label consistency. The resulting stable part assignments are used in the subsequent parameter optimization. Detailed implementation of the full pipeline is provided in the supplementary material.

\noindent \textbf{Physical Parameter Optimization.}
Given the part-level assignments, we associate each part with a learnable physical parameter vector. Suppose the scene is decomposed into $M$ parts, and each part is assigned a physical parameter vector. In our implementation, these parameters include both constitutive parameters and material attributes, such as Young's modulus, Poisson's ratio, yield stress, plastic viscosity, internal friction angle, and density. Let $\boldsymbol{\theta}_m \in \mathbb{R}_{+}^{K}$ denote the physical parameter vector of the $m$-th part, where $K$ is the number of optimized physical quantities, and let $\boldsymbol{\theta}_m^{*}$ denote its initialized value obtained from the simulation-driven initialization stage.

To enforce positivity and improve optimization stability, we optimize the parameters in the log domain. Specifically, we introduce an unconstrained trainable variable $\boldsymbol{\xi}_m \in \mathbb{R}^{K}$ for each part and initialize it as
\begin{equation}
    \boldsymbol{\xi}_m^{(0)} = \log\!\bigl(\max(\boldsymbol{\theta}_m^{*}, \varepsilon)\bigr),
    \label{eq:log_init_vec}
\end{equation}
where $\log(\cdot)$ and $\max(\cdot,\varepsilon)$ are applied element-wise, and $\varepsilon$ is a small positive constant for numerical stability. During optimization, we first clamp $\boldsymbol{\xi}_m$ to the feasible log-domain range and then recover the physical parameters by
\begin{equation}
    \boldsymbol{\xi}_m \leftarrow \operatorname{clip}\!\left(
    \boldsymbol{\xi}_m,\;
    \log\!\bigl(\max(\mathbf{l},\varepsilon)\bigr),\;
    \log\!\bigl(\max(\mathbf{u},\varepsilon)\bigr)
    \right),
    \label{eq:log_clip_vec}
\end{equation}
\begin{equation}
    \boldsymbol{\theta}_m = \exp(\boldsymbol{\xi}_m),
    \label{eq:log_recover_vec}
\end{equation}
where $\mathbf{l}, \mathbf{u} \in \mathbb{R}^{K}$ denote the lower and upper bounds of the physical parameters, respectively, and $\exp(\cdot)$ is also applied element-wise.

The resulting part-level parameters are assigned to all particles belonging to the corresponding part and are then used by the differentiable MPM simulator. Starting from the initialized parameters, we unroll the MPM simulator and differentiable renderer, and optimize the log-domain variables by minimizing the dual-domain motion supervision:
\begin{equation}
    \min_{\{\boldsymbol{\xi}_m\}_{m=1}^{M}}
    \mathcal{L}_{\mathrm{DMS}}\bigl(\{\boldsymbol{\theta}_m\}_{m=1}^{M}\bigr),
    \qquad
    \boldsymbol{\theta}_m = \exp(\boldsymbol{\xi}_m).
    \label{eq:final_opt_xi}
\end{equation}
After the simulator computes gradients with respect to the physical parameters, the gradients are propagated back to the log-domain variables through the chain rule
\begin{equation}
    \frac{\partial \mathcal{L}}{\partial \boldsymbol{\xi}_m}
    =
    \frac{\partial \mathcal{L}}{\partial \boldsymbol{\theta}_m}
    \odot
    \boldsymbol{\theta}_m,
    \label{eq:log_grad_vec}
\end{equation}
where $\odot$ denotes element-wise multiplication. The trainable log-domain variables $\{\boldsymbol{\xi}_m\}_{m=1}^{M}$ are then updated using AdamW\cite{loshchilov2017decoupled}. At each optimization step, the current $\boldsymbol{\xi}_m$ is first mapped to the physical parameters through Eq.~\eqref{eq:log_recover_vec}, and the resulting gradients are back-propagated to $\boldsymbol{\xi}_m$ for parameter update. For additional stability, we clip the gradients of the trainable log-domain variables during optimization.

In summary, the optimization proceeds as:
\[
    \{\boldsymbol{\theta}_m^{*}\}_{m=1}^{M}
    \rightarrow
    \{\boldsymbol{\xi}_m^{(0)}\}_{m=1}^{M}
    \rightarrow
    \{\boldsymbol{\theta}_m\}_{m=1}^{M}
    \rightarrow
    \hat{\mathbf{V}}
    \rightarrow
    \mathcal{L}_{\mathrm{DMS}}(\hat{\mathbf{V}}, \mathbf{V}^{*}),
\]
Starting from the initialized part-level parameters $\{\boldsymbol{\theta}_m^{*}\}_{m=1}^{M}$, we initialize the log-domain variables $\{\boldsymbol{\xi}_m^{(0)}\}_{m=1}^{M}$, recover the physical parameters $\{\boldsymbol{\theta}_m\}_{m=1}^{M}$, and optimize them through differentiable simulation and rendering under the dual-domain motion supervision loss $\mathcal{L}_{\mathrm{DMS}}(\hat{\mathbf{V}}, \mathbf{V}^{*})$. Overall, this formulation enables stable optimization of part-level physical parameters under limited-view visual supervision.
\section{Experiments}

\begin{table*}[t]
    \centering
    \caption{Quantitative comparison on four scenes from the PhysDreamer dataset.
        We report video fidelity metrics, including MS-SSIM, PSNR, and LPIPS, together with peak GPU memory usage.
        The column ``all'' denotes the average results over all four scenes.
        Best  results are highlighted in bold, respectively.
    }
    \label{tab:exp_exp1}
    \resizebox{\textwidth}{!}{
        \begin{tabular}{llcccccccccccccccccccc}
            \hline
            \multirow{2}{*}{Method}         & \multirow{2}{*}{Proc.\ \& Year}   & \multicolumn{4}{c}{Alocasia}  &
            \multicolumn{4}{c}{Carnations}  & \multicolumn{4}{c}{Hat}           & \multicolumn{4}{c}{Telephone} & \multicolumn{4}{c}{All} \\
            \cmidrule(lr){3-6} \cmidrule(lr){7-10} \cmidrule(lr){11-14} \cmidrule(lr){15-18} \cmidrule(lr){19-22}
            {}  &{} &    
            MS-SSIM$\uparrow$ & PSNR$\uparrow$ & LPIPS$\downarrow$ & Memory$\downarrow$ &
            MS-SSIM$\uparrow$ & PSNR$\uparrow$ & LPIPS$\downarrow$ & Memory$\downarrow$ &
            MS-SSIM$\uparrow$ & PSNR$\uparrow$ & LPIPS$\downarrow$ & Memory$\downarrow$ &
            MS-SSIM$\uparrow$ & PSNR$\uparrow$ & LPIPS$\downarrow$ & Memory$\downarrow$ &
            MS-SSIM$\uparrow$ & PSNR$\uparrow$ & LPIPS$\downarrow$ & Memory$\downarrow$ \\
            \hline
            DreamPhysics    &   AAAI2025    &
            \textbf{0.8143} &   \textbf{18.52}  &   \textbf{0.3904} &   32.21   &
            0.7458          &   17.55           &   0.2199          &   34.12   &
            0.5068          &   14.69           &   0.2118          &   33.06   &   
            0.7886          &   21.18           &   0.5117          &   34.27   &
            0.7139          &   17.98           &   0.3334          &   33.42   \\
            
            Physics3D       & \multicolumn{1}{c}{-}     &
            0.6807          &   16.89           &   0.4236          &   31.53   &   
            0.6715          &   17.05           &   0.1964          &   33.91   &
            0.5057          &   14.75           &   0.1970          &   32.85   &   
            0.7929          &   20.94           &   0.4011          &   33.94   &
            0.6627          &   17.41           &   0.3046          &   33.06   \\
            
            PhysFlow&   CVPR2025&   
            0.5785          &   15.60           &   0.4874          &   38.38   &   
            0.6448          &   16.03           &   0.2460          &   38.53   &
            0.5091          &   14.76           &   0.2004          &   36.68   &   
            0.7996          &   21.07           &   0.3943          &   40.48   &
            0.6330          &   16.87           &   0.3320          &   38.52   \\
            
            Ours            &   \multicolumn{1}{c}{-}   &
            0.7958          &   18.40           &   0.3958          &   \textbf{22.32}  &
            \textbf{0.7900} &   \textbf{19.21}  &   \textbf{0.1257} &   \textbf{22.14}  &   
            \textbf{0.5109} &   \textbf{14.80}  &   \textbf{0.1966} &   \textbf{21.65}  &
            \textbf{0.8325} &   \textbf{21.82}  &   \textbf{0.4246} &   \textbf{20.97}  &   
            \textbf{0.7323} &   \textbf{18.56}  &   \textbf{0.2857} &   \textbf{21.77}  \\
            \hline
        \end{tabular}
        }
\end{table*}

\begin{figure*}[t]
    \centering
    \includegraphics[width=0.8\textwidth]{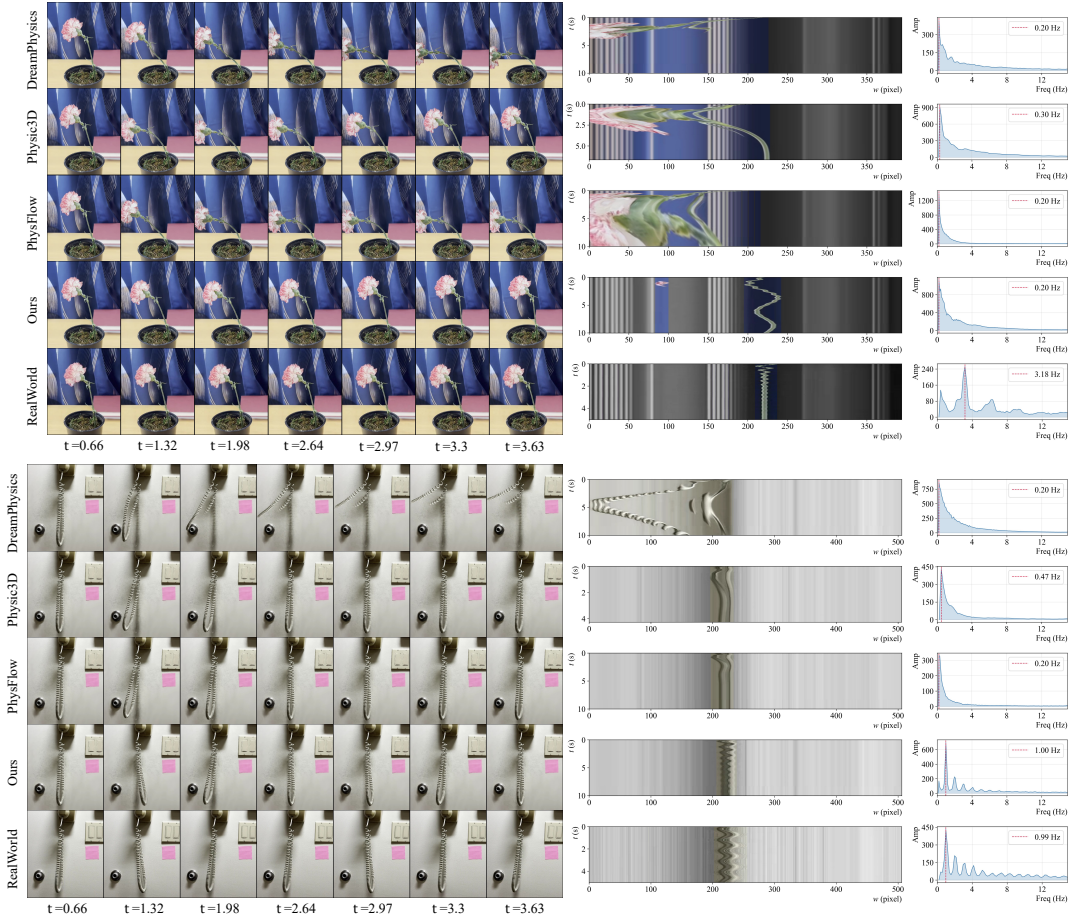}
    \caption{
        Visual comparison on four scenes from the PhysDreamer dataset.
        For each method, we show representative video frames, the motion-region ww-tt visualization, and the corresponding FFT spectrum.
        The $w$-$t$ maps reveal temporal motion patterns along a fixed spatial coordinate, while the FFT plots further characterize their frequency responses.
        In the \emph{carnations} scene, our method better preserves the oscillatory motion property, producing clearer periodic patterns and frequency peaks that are closer to the real-world reference.
        In the \emph{telephone} scene, our method achieves the closest alignment with the real-world motion in both the spatiotemporal trajectory and the spectral distribution.
    }
    \Description{The figure is organized into an upper group and a lower group, each corresponding to one scene. In each group, five rows are shown for DreamPhysics, Physics3D, PhysFlow, Ours, and RealWorld. For every row, the left section contains a sequence of video frames sampled at multiple time steps. The middle section shows a corresponding $w$-$t$ spatiotemporal slice extracted along a fixed spatial coordinate, with horizontal position on the horizontal axis and time on the vertical axis. The right section shows the corresponding FFT spectrum, with frequency on the horizontal axis and amplitude on the vertical axis. The upper group depicts a flower scene, and the lower group depicts a hanging telephone scene.}
    \label{fig:result1}
\end{figure*}

In this section, we evaluate Resonance4D on both real-world and synthetic benchmarks. We first present results on real data to assess its ability to capture realistic motion and physical behaviors from visual observations (Sec.~4.2). We then perform further evaluation on synthetic scenes, which enables more comprehensive qualitative and quantitative comparisons (Sec.~4.3). Finally, we conduct ablation studies to analyze the effectiveness of the key design choices in our framework (Sec.~4.4). More visualizations and detailed analyses are provided in the appendix.

\subsection{Experiments Details}

\noindent \textbf{Datasets} We conduct experiments on both real-world and synthetic datasets. For real-world evaluation, we adopt four scenes from the PhysDreamer Dataset\cite{zhang2024physdreamer}: \textit{telephone}, \textit{carnations}, \textit{alocasia}, and \textit{hat}. 
Each scene contains a captured motion video together with its corresponding reconstructed 3D Gaussian representation, providing a suitable testbed for evaluating physics-driven parameter optimization under realistic appearance variations and complex motion dynamics. 
For synthetic evaluation, we use four scenes from the PAC-NeRF Dataset\cite{li2023pac}, including \textit{bird}, \textit{cat}, and \textit{toothpaste}. In addition to rendered observations, PAC-NeRF provides simulated point-wise motion trajectories, enabling more detailed and systematic evaluation of recovered dynamic behaviors.

\noindent \textbf{Metrics}
For real-world evaluation, we measure the discrepancy between the rendered videos and the captured videos using MS-SSIM\cite{wang2003multiscale}, PSNR, and LPIPS\cite{zhang2018unreasonable}. These metrics assess reconstruction quality from structural, pixel-wise, and perceptual perspectives, respectively.
For synthetic evaluation, we compare the simulated moving points with the ground-truth trajectories using Chamfer Distance (CD), 95th percentile Hausdorff Distance (HD95), and F-score to assess the geometric accuracy of the recovered dynamics.

\noindent \textbf{Baselines}
For the PhysDreamer dataset, we compare our method with representative approaches that are compatible with our experimental setting and provide complete training pipelines, including DreamPhysics, Physics3D, and PhysFlow. These baselines reflect different levels of physical modeling complexity: DreamPhysics focuses on optimizing a single physical property, Physics3D supports joint optimization of multiple physical parameters, and PhysFlow further considers multi-material scenes with multiple physical properties. A more detailed comparison is provided in Tab.\ref{tab:intro_support_phys}.
For the PAC-NeRF dataset, we use PAC-NeRF and PhysFlow as baselines.
We exclude GIC\cite{cai2024gic} and MASIV\cite{zhao2025toward} from the comparison, since both methods further refine the 3D Gaussian representation after physical parameter optimization. This additional geometry optimization changes the problem setting and therefore prevents a strictly fair comparison with our target task.

\noindent \textbf{Implementation Details}
We use the official implementations of all baseline methods and follow their default training settings whenever applicable. For fair comparison, all rendered videos are exported at 30 FPS during evaluation, and all experiments are performed on a single NVIDIA L20 GPU with 48GB memory. For both the PhysDreamer and PAC-NeRF datasets, we use the reconstructed 3D Gaussian scenes released by PhysFlow. For PhysDreamer dataset, we further use the manually separated moving parts provided in the same release.

\subsection{Results on the PhysDreamer Real-World Dataset}
We first evaluate Resonance4D on the PhysDreamer real-world benchmark, which contains four challenging scenes with diverse motion patterns and appearance variations. Quantitative results are reported in Table~\ref{tab:exp_exp1}, where we compare with representative physics-driven 3DGS baselines using PSNR, MS-SSIM, LPIPS, and peak GPU memory.

Overall, Resonance4D delivers the strongest overall performance across all four scenes, achieving 0.7323 MS-SSIM, 18.56 PSNR, and 0.2857 LPIPS, with only 21.77,GB peak GPU memory. Compared with DreamPhysics, the most competitive baseline, it improves the average PSNR by 0.58,dB and reduces LPIPS from 0.3334 to 0.2857, while using substantially less memory during optimization. This suggests that high-quality dynamic reconstruction does not require heavy online priors: lightweight dual-domain motion supervision can already provide effective guidance at a much lower computational cost. Resonance4D also consistently outperforms Physics3D and PhysFlow in both reconstruction quality and efficiency, demonstrating a more favorable balance between fidelity and optimization overhead.

Resonance4D shows the clearest advantage on \textit{carnations} and \textit{telephone}. On \textit{carnations}, it achieves the best results on all three image-based metrics, and the recovered motion is more closely aligned with the ground-truth temporal behavior. On \textit{telephone}, it attains the highest MS-SSIM and PSNR, while the dominant frequency peak is closer to that of the real-world reference. Together, these results suggest that Resonance4D improves not only frame-wise appearance, but also the fidelity of underlying motion dynamics.

Overall, the results on the PhysDreamer benchmark show that lightweight dual-domain motion supervision is sufficient to recover realistic and physically plausible motion in real-world scenes, while significantly reducing the optimization memory cost compared with prior methods.

\subsection{Results on the PAC-NeRF Synthetic Dataset}

Table\ref{tab:exp_exp2} reports the quantitative results on the PAC-NeRF benchmark. Overall, our method achieves competitive performance against representative physics-driven baselines. 
Specifically, it obtains the best CD and HD95, improving them to 0.0063 and 0.0914, respectively, which indicates more accurate geometric alignment and smaller worst-case deviation from the ground truth. In terms of F-score, our method reaches 0.4256, which is slightly lower than PAC-NeRF but remains at a comparable level. These results suggest that our method is broadly on par with mainstream approaches on this benchmark, while showing an advantage in geometric accuracy.

Figure~\ref{fig:result2} further provides qualitative comparisons on the \textit{bird} and \textit{toothpaste} scenes at consecutive time steps. 
In the relatively stiff \textit{bird} scene, our results remain closer to the ground truth in overall shape evolution, while PAC-NeRF tends to produce overly soft deformation and PhysFlow exhibits a somewhat rigid response. 
In the softer \textit{toothpaste} scene, our method also better matches the ground-truth temporal evolution, whereas PhysFlow shows an unnatural rebound in shape change that is inconsistent with the reference. 
Combined with the quantitative results, these visualizations indicate that our method can recover material-dependent dynamics in a stable and plausible manner, yielding competitive performance across different motion regimes.

Overall, our method remains competitive with mainstream approaches on PAC-NeRF.

\begin{figure}[t]
    \centering
    \includegraphics[width=\linewidth]{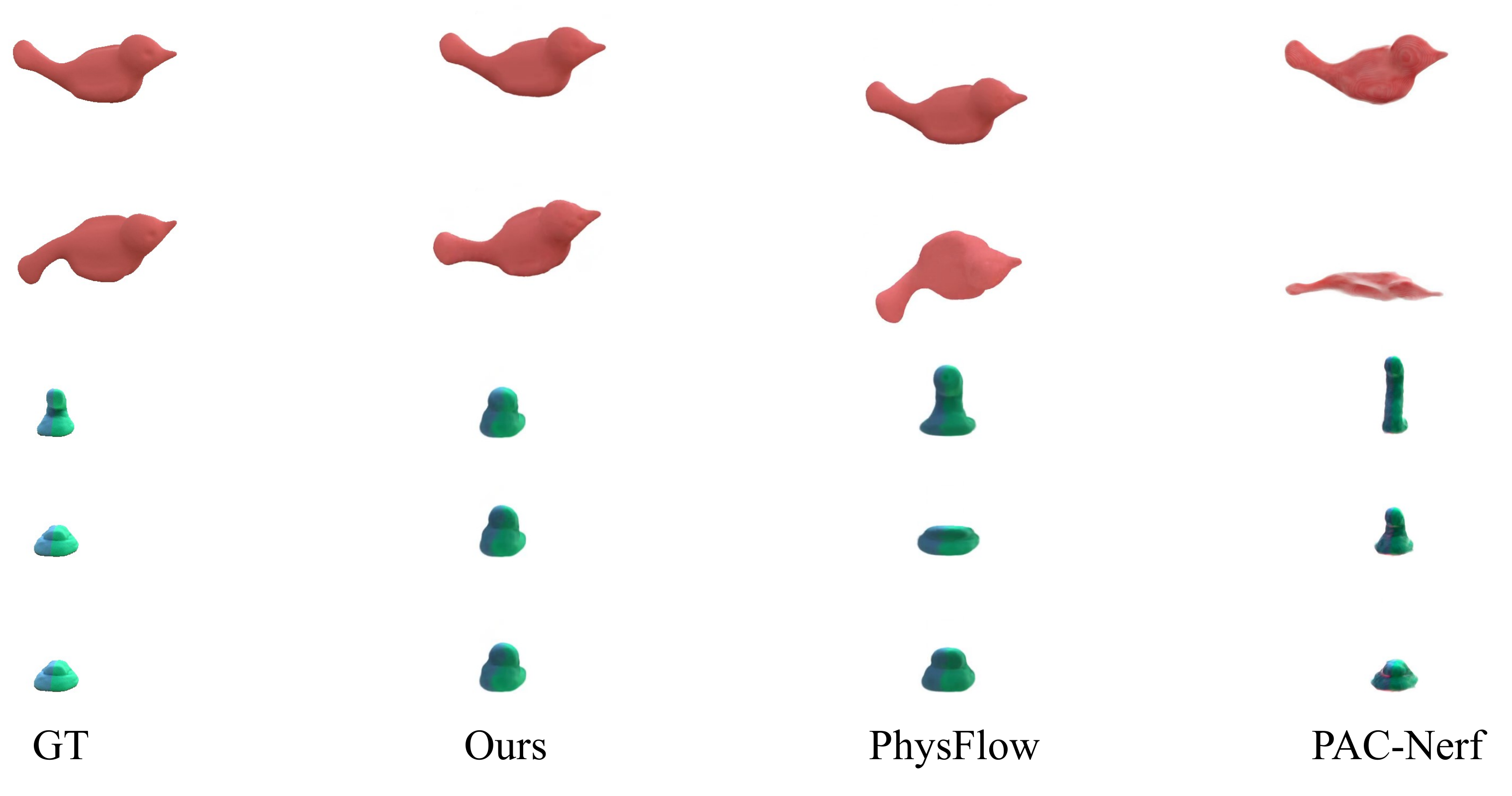}
    \caption{
    Qualitative comparison on the \textit{bird} and \textit{toothpaste} scenes at consecutive time steps.
    The \textit{bird} scene corresponds to relatively stiff material behavior, while \textit{toothpaste} exhibits softer dynamics.
    Our method better matches the material-dependent motion in both cases.
    PAC-NeRF tends to produce overly soft deformation in the stiff \textit{bird} scene, whereas PhysFlow is overly rigid.
    In the \textit{toothpaste} scene, PhysFlow shows an unnatural rebound in shape evolution, which deviates from the ground truth.
    }
    \Description{The figure presents a qualitative comparison on two scenes across consecutive time steps. Four columns are shown for GT, Ours, PhysFlow, and PAC-NeRF. In each column, the upper row group shows a sequence from the bird scene, and the lower row group shows a sequence from the toothpaste scene. Each sequence contains multiple shapes sampled at successive times, arranged from top to bottom. The bird shapes are shown in red, and the toothpaste shapes are shown in green to blue.}
    \label{fig:result2}
\end{figure}

\begin{table}[t]
    \centering
    \footnotesize
    \caption{Quantitative comparison on the PAC-NeRF dataset. We report CD, HD95, and F-score. Best results are highlighted in bold.}
    \label{tab:exp_exp2}
        \begin{tabular}{lcccc}
            \toprule
            Method & Proc.\ \& Year & CD$\downarrow$ & HD95$\downarrow$ & F-score$\uparrow$ \\
            \midrule
            
            PAC-NeRF
            & ICLR 2023
            &0.0074 	&0.1013 	&\textbf{0.4329}   \\

            PhysFlow
            & CVPR 2025
            &0.0085 	&0.1113 	&0.3647      \\

            Ours
            & -
            &\textbf{0.0063} 	&\textbf{0.0914} 	&0.4256  \\

            \bottomrule
        \end{tabular}
\end{table}

\subsection{Ablation}
We perform ablation studies on four scenes from the PhysDreamer dataset to examine the effectiveness of the major components in our framework. All ablation variants are evaluated under a unified setting using three image-space metrics: MS-SSIM, PSNR, and LPIPS. The study covers four aspects, including Dual-domain Motion Supervision, simulation-driven initialization, parameter update granularity, and density optimization. Additional results on moving-part extraction are deferred to the appendix.

\noindent \textbf{Ablation on Dual-domain Motion Supervision.}
We evaluate different combinations of spatial-domain and frequency-domain supervision, including L1, SSIM, and FFT-based spectral supervision. The results are reported in Table~\ref{tab:exp_abs_loss}. Using only spatial supervision already produces reasonable reconstruction quality, while replacing L1 with SSIM brings consistent improvements, indicating that structural similarity provides a stronger appearance-level constraint. Further introducing FFT-based frequency-domain supervision leads to the best performance on all three metrics. This suggests that spectral supervision provides complementary motion cues beyond spatial appearance matching, and helps better capture local dynamic patterns. Overall, these results verify that the proposed dual-domain motion supervision achieves a more effective balance between appearance alignment and motion modeling.
\begin{table}[t]
    \centering
    \footnotesize
    \caption{Ablation study on dual-domain motion supervision. We compare different combinations of spatial-domain and frequency-domain supervision.}
    \label{tab:exp_abs_loss}
    \renewcommand{\arraystretch}{1.1}
    \resizebox{\linewidth}{!}{
    \begin{tabular}{cccccc}
        \hline
        \multicolumn{2}{c}{Spatial Domain} & \multicolumn{1}{c}{Frequency Domain} & \multicolumn{3}{c}{Metrics} \\
        \cmidrule(lr){1-2} \cmidrule(lr){3-3} \cmidrule(lr){4-6}
        L1 & SSIM & FFT & MS-SSIM$\uparrow$ & PSNR$\uparrow$ & LPIPS$\downarrow$ \\
        \hline
        $\checkmark$ &              &               & 0.7268 & 18.4173 & 0.2913 \\
                     & $\checkmark$ &               & 0.7284 & 18.4463 & 0.2899 \\
        $\checkmark$ &              & $\checkmark$  & 0.7304 & 18.5486 & 0.2885 \\
                     & $\checkmark$ & $\checkmark$  & \textbf{0.7323} & \textbf{18.5570} & \textbf{0.2857} \\
        \hline
    \end{tabular}
    }
\end{table}

\noindent \textbf{Ablation on Simulation-driven Initialization.}
We evaluate the proposed simulation-driven initialization. Because joint optimization of coupled physical parameters is highly sensitive to initialization, poor starting points can lead to unstable convergence or inferior local optima. Table~\ref{tab:exp_abs_init} shows that simulation-driven initialization consistently improves the final results over simpler baselines, indicating that coarse candidate selection provides a useful prior for subsequent optimization. Although it introduces extra offline cost, this cost is incurred only once and remains lighter than relying on heavy online teachers throughout training.
\begin{table}[t]
    \centering
    \footnotesize
    \caption{Ablation study on initialization strategies. We compare GPT-based initialization and simulation-driven initialization, both before and after optimization, and also report the average time cost of simulation-driven initialization.}
    \label{tab:exp_abs_init}
    \begin{tabular}{lccc}
        \toprule
        Method                     & MS-SSIM$\uparrow$ & PSNR$\uparrow$ & LPIPS$\downarrow$ \\
        \midrule
        GPT Init.                  & 0.6273            & 16.8943        & 0.3866            \\
        GPT Init. + Training       & 0.6272            & 16.8653        & 0.3852            \\
        Sampling Init.             & 0.7296            & 18.4854        & 0.2886            \\
        Sampling Init. + Training  & 0.7323            & 18.5570        & 0.2857            \\
        \midrule
        \multicolumn{4}{l}{\textbf{Avg. sampling init. time}: 14.0 min} \\
        \multicolumn{4}{l}{\hspace{1.5em}Four scenes, 1080P, 32 videos, 16 frames.} \\
        \bottomrule
    \end{tabular}
\end{table}

\noindent \textbf{Ablation on Parameter Update Granularity.}
We compare object-level, particle-level, and part-level parameter optimization. Table~\ref{tab:exp_abs_update} shows that object-level updates are too coarse for heterogeneous materials, while particle-level updates are unstable under limited supervision. Part-level optimization achieves the best overall performance, providing a better balance between expressiveness and stability.
\begin{table}[t]
    \centering
    \footnotesize
    \caption{Ablation study on parameter update granularity. We compare particle-level, object-level, and part-level updates.}
    \label{tab:exp_abs_update}
    \begin{tabular}{lccc}
        \hline
                          & MS-SSIM$\uparrow$ & PSNR$\uparrow$ & LPIPS$\downarrow$ \\
        \hline
        Particle Update   &    0.7241       &   18.2771 &    0.2935 \\
        Object Update     &    0.7269       &   18.4318 &    0.2909 \\
        Part Update       &    0.7323       &   18.5570 &    0.2857 \\
        \hline
    \end{tabular}
\end{table}

\noindent \textbf{Ablation on Density Optimization.} A key difference between Resonance4D and several previous methods is that we jointly optimize not only constitutive parameters but also material attributes such as density. To evaluate the impact of density optimization, we compare variants with and without density in Table~\ref{tab:exp_abs_density}. The results in Table~\ref{tab:exp_abs_density} show that optimizing density leads to consistent improvements. This suggests that fixing density as a constant can limit the expressiveness of the physical model, especially when the target dynamics depend on both constitutive behavior and mass-related properties. Jointly optimizing density therefore improves the completeness of parameter fitting and leads to more accurate dynamic reconstruction.
\begin{table}[t]
    \centering
    \footnotesize
    \caption{Effect of density optimization in the proposed framework.}
    \label{tab:exp_abs_density}
    \begin{tabular}{lccc}
        \hline
                            & MS-SSIM$\uparrow$ & PSNR$\uparrow$ & LPIPS$\downarrow$ \\
        \hline
        W/O density Update      & 0.7099 & 17.9778 & 0.3020 \\
        With density Update     & 0.7323 & 18.5570 & 0.2857 \\
        \hline
    \end{tabular}
\end{table}
\section{Conclusion}

In this paper, we presented a physics-driven dynamic generation framework based on 3D Gaussians for recovering part-level physical parameters from visual observations and synthesizing plausible motion. Through dual-domain motion supervision, our method provides a more direct and lightweight dynamic constraint for physical parameter optimization, in contrast to heavy online motion priors. 
Experimental results show that the proposed approach achieves more stable parameter recovery and produces more natural dynamic motion. The current framework still relies on relatively complete reconstruction and lacks high-level feedback from vision-language models; future work will extend it to sparse-view settings and incorporate vision-language model feedback to further improve dynamic modeling performance.

\begin{acks}
To Robert, for the bagels and explaining CMYK and color spaces.
\end{acks}

\bibliographystyle{ACM-Reference-Format}
\bibliography{inference}

@String{Computer = "{IEEE} Computer" }

@String{Springer = "Springer-Verlag" }

@article{abou2024physically,
  title={Physically embodied gaussian splatting: A realtime correctable world model for robotics},
  author={Abou-Chakra, Jad and Rana, Krishan and Dayoub, Feras and S{\"u}nderhauf, Niko},
  journal={arXiv preprint arXiv:2406.10788},
  year={2024}
}

@inproceedings{tseng2025gaussian,
  title={Gaussian Splatting Visual MPC for Granular Media Manipulation},
  author={Tseng, Wei-Cheng and Zhang, Ellina and Jatavallabhula, Krishna Murthy and Shkurti, Florian},
  booktitle={2025 IEEE International Conference on Robotics and Automation (ICRA)},
  pages={3176--3182},
  year={2025},
  organization={IEEE}
}

@article{jiang2024robust,
  title={Robust dual gaussian splatting for immersive human-centric volumetric videos},
  author={Jiang, Yuheng and Shen, Zhehao and Hong, Yu and Guo, Chengcheng and Wu, Yize and Zhang, Yingliang and Yu, Jingyi and Xu, Lan},
  journal={ACM Transactions on Graphics (TOG)},
  volume={43},
  number={6},
  pages={1--15},
  year={2024},
  publisher={ACM New York, NY, USA}
}

@article{jin2024pyramidal,
  title={Pyramidal flow matching for efficient video generative modeling},
  author={Jin, Yang and Sun, Zhicheng and Li, Ningyuan and Xu, Kun and Jiang, Hao and Zhuang, Nan and Huang, Quzhe and Song, Yang and Mu, Yadong and Lin, Zhouchen},
  journal={arXiv preprint arXiv:2410.05954},
  year={2024}
}

@article{kerbl20233d,
  title={3d gaussian splatting for real-time radiance field rendering.},
  author={Kerbl, Bernhard and Kopanas, Georgios and Leimk{\"u}hler, Thomas and Drettakis, George and others},
  journal={ACM Trans. Graph.},
  volume={42},
  number={4},
  pages={139--1},
  year={2023}
}

@article{hu2018moving,
  title={A moving least squares material point method with displacement discontinuity and two-way rigid body coupling},
  author={Hu, Yuanming and Fang, Yu and Ge, Ziheng and Qu, Ziyin and Zhu, Yixin and Pradhana, Andre and Jiang, Chenfanfu},
  journal={ACM Transactions on Graphics (TOG)},
  volume={37},
  number={4},
  pages={1--14},
  year={2018},
  publisher={ACM New York, NY, USA}
}

@inproceedings{zhang2024physdreamer,
  title={Physdreamer: Physics-based interaction with 3d objects via video generation},
  author={Zhang, Tianyuan and Yu, Hong-Xing and Wu, Rundi and Feng, Brandon Y and Zheng, Changxi and Snavely, Noah and Wu, Jiajun and Freeman, William T},
  booktitle={European Conference on Computer Vision},
  pages={388--406},
  year={2024},
  organization={Springer}
}

@inproceedings{huang2025dreamphysics,
  title={Dreamphysics: Learning physics-based 3d dynamics with video diffusion priors},
  author={Huang, Tianyu and Zhang, Haoze and Zeng, Yihan and Zhang, Zhilu and Li, Hui and Zuo, Wangmeng and Lau, Rynson WH},
  booktitle={Proceedings of the AAAI Conference on Artificial Intelligence},
  volume={39},
  number={4},
  pages={3733--3741},
  year={2025}
}

@article{liu2024physics3d,
  title={Physics3d: Learning physical properties of 3d gaussians via video diffusion},
  author={Liu, Fangfu and Wang, Hanyang and Yao, Shunyu and Zhang, Shengjun and Zhou, Jie and Duan, Yueqi},
  journal={arXiv preprint arXiv:2406.04338},
  year={2024}
}

@inproceedings{xie2024physgaussian,
  title={Physgaussian: Physics-integrated 3d gaussians for generative dynamics},
  author={Xie, Tianyi and Zong, Zeshun and Qiu, Yuxing and Li, Xuan and Feng, Yutao and Yang, Yin and Jiang, Chenfanfu},
  booktitle={Proceedings of the IEEE/CVF Conference on Computer Vision and Pattern Recognition},
  pages={4389--4398},
  year={2024}
}

@article{lin2025omniphysgs,
  title={Omniphysgs: 3d constitutive gaussians for general physics-based dynamics generation},
  author={Lin, Yuchen and Lin, Chenguo and Xu, Jianjin and Mu, Yadong},
  journal={arXiv preprint arXiv:2501.18982},
  year={2025}
}

@inproceedings{liu2025unleashing,
  title={Unleashing the potential of multi-modal foundation models and video diffusion for 4d dynamic physical scene simulation},
  author={Liu, Zhuoman and Ye, Weicai and Luximon, Yan and Wan, Pengfei and Zhang, Di},
  booktitle={Proceedings of the Computer Vision and Pattern Recognition Conference},
  pages={11016--11025},
  year={2025}
}

@article{mildenhall2021nerf,
  title={Nerf: Representing scenes as neural radiance fields for view synthesis},
  author={Mildenhall, Ben and Srinivasan, Pratul P and Tancik, Matthew and Barron, Jonathan T and Ramamoorthi, Ravi and Ng, Ren},
  journal={Communications of the ACM},
  volume={65},
  number={1},
  pages={99--106},
  year={2021},
  publisher={ACM New York, NY, USA}
}

@inproceedings{pumarola2021d,
  title={D-nerf: Neural radiance fields for dynamic scenes},
  author={Pumarola, Albert and Corona, Enric and Pons-Moll, Gerard and Moreno-Noguer, Francesc},
  booktitle={Proceedings of the IEEE/CVF conference on computer vision and pattern recognition},
  pages={10318--10327},
  year={2021}
}

@inproceedings{park2021nerfies,
  title={Nerfies: Deformable neural radiance fields},
  author={Park, Keunhong and Sinha, Utkarsh and Barron, Jonathan T and Bouaziz, Sofien and Goldman, Dan B and Seitz, Steven M and Martin-Brualla, Ricardo},
  booktitle={Proceedings of the IEEE/CVF international conference on computer vision},
  pages={5865--5874},
  year={2021}
}

@article{park2021hypernerf,
  title={Hypernerf: A higher-dimensional representation for topologically varying neural radiance fields},
  author={Park, Keunhong and Sinha, Utkarsh and Hedman, Peter and Barron, Jonathan T and Bouaziz, Sofien and Goldman, Dan B and Martin-Brualla, Ricardo and Seitz, Steven M},
  journal={arXiv preprint arXiv:2106.13228},
  year={2021}
}

@inproceedings{luiten2024dynamic,
  title={Dynamic 3d gaussians: Tracking by persistent dynamic view synthesis},
  author={Luiten, Jonathon and Kopanas, Georgios and Leibe, Bastian and Ramanan, Deva},
  booktitle={2024 International Conference on 3D Vision (3DV)},
  pages={800--809},
  year={2024},
  organization={IEEE}
}

@inproceedings{wu20244d,
  title={4d gaussian splatting for real-time dynamic scene rendering},
  author={Wu, Guanjun and Yi, Taoran and Fang, Jiemin and Xie, Lingxi and Zhang, Xiaopeng and Wei, Wei and Liu, Wenyu and Tian, Qi and Wang, Xinggang},
  booktitle={Proceedings of the IEEE/CVF conference on computer vision and pattern recognition},
  pages={20310--20320},
  year={2024}
}

@article{blattmann2023stable,
  title={Stable video diffusion: Scaling latent video diffusion models to large datasets},
  author={Blattmann, Andreas and Dockhorn, Tim and Kulal, Sumith and Mendelevitch, Daniel and Kilian, Maciej and Lorenz, Dominik and Levi, Yam and English, Zion and Voleti, Vikram and Letts, Adam and others},
  journal={arXiv preprint arXiv:2311.15127},
  year={2023}
}

@inproceedings{chen2024videocrafter2,
  title={Videocrafter2: Overcoming data limitations for high-quality video diffusion models},
  author={Chen, Haoxin and Zhang, Yong and Cun, Xiaodong and Xia, Menghan and Wang, Xintao and Weng, Chao and Shan, Ying},
  booktitle={Proceedings of the IEEE/CVF conference on computer vision and pattern recognition},
  pages={7310--7320},
  year={2024}
}

@article{yang2024cogvideox,
  title={Cogvideox: Text-to-video diffusion models with an expert transformer},
  author={Yang, Zhuoyi and Teng, Jiayan and Zheng, Wendi and Ding, Ming and Huang, Shiyu and Xu, Jiazheng and Yang, Yuanming and Hong, Wenyi and Zhang, Xiaohan and Feng, Guanyu and others},
  journal={arXiv preprint arXiv:2408.06072},
  year={2024}
}

@article{liu2024sora,
  title={Sora: A review on background, technology, limitations, and opportunities of large vision models},
  author={Liu, Yixin and Zhang, Kai and Li, Yuan and Yan, Zhiling and Gao, Chujie and Chen, Ruoxi and Yuan, Zhengqing and Huang, Yue and Sun, Hanchi and Gao, Jianfeng and others},
  journal={arXiv preprint arXiv:2402.17177},
  year={2024}
}

@article{gu2025long,
  title={Long-context autoregressive video modeling with next-frame prediction},
  author={Gu, Yuchao and Mao, Weijia and Shou, Mike Zheng},
  journal={arXiv preprint arXiv:2503.19325},
  year={2025}
}

@article{yang2025longlive,
  title={Longlive: Real-time interactive long video generation},
  author={Yang, Shuai and Huang, Wei and Chu, Ruihang and Xiao, Yicheng and Zhao, Yuyang and Wang, Xianbang and Li, Muyang and Xie, Enze and Chen, Yingcong and Lu, Yao and others},
  journal={arXiv preprint arXiv:2509.22622},
  year={2025}
}

@inproceedings{zhao2025physsplat,
  title={PhysSplat: Efficient Physics Simulation for 3D Scenes via MLLM-Guided Gaussian Splatting},
  author={Zhao, Haoyu and Wang, Hao and Zhao, Xingyue and Fei, Hao and Wang, Hongqiu and Long, Chengjiang and Zou, Hua},
  booktitle={Proceedings of the IEEE/CVF International Conference on Computer Vision},
  pages={5242--5252},
  year={2025}
}

@article{dinov3,
  title={Dinov3},
  author={Sim{\'e}oni, Oriane and Vo, Huy V and Seitzer, Maximilian and Baldassarre, Federico and Oquab, Maxime and Jose, Cijo and Khalidov, Vasil and Szafraniec, Marc and Yi, Seungeun and Ramamonjisoa, Micha{\"e}l and others},
  journal={arXiv preprint arXiv:2508.10104},
  year={2025}
}

@article{sam3,
  title={Sam 3: Segment anything with concepts},
  author={Carion, Nicolas and Gustafson, Laura and Hu, Yuan-Ting and Debnath, Shoubhik and Hu, Ronghang and Suris, Didac and Ryali, Chaitanya and Alwala, Kalyan Vasudev and Khedr, Haitham and Huang, Andrew and others},
  journal={arXiv preprint arXiv:2511.16719},
  year={2025}
}

@inproceedings{marrie2025ludvig,
  title={Ludvig: Learning-free uplifting of 2d visual features to gaussian splatting scenes},
  author={Marrie, Juliette and M{\'e}n{\'e}gaux, Romain and Arbel, Michael and Larlus, Diane and Mairal, Julien},
  booktitle={Proceedings of the IEEE/CVF International Conference on Computer Vision},
  pages={7440--7450},
  year={2025}
}

@article{loshchilov2017decoupled,
  title={Decoupled weight decay regularization},
  author={Loshchilov, Ilya and Hutter, Frank},
  journal={arXiv preprint arXiv:1711.05101},
  year={2017}
}

@article{li2023pac,
  title={Pac-nerf: Physics augmented continuum neural radiance fields for geometry-agnostic system identification},
  author={Li, Xuan and Qiao, Yi-Ling and Chen, Peter Yichen and Jatavallabhula, Krishna Murthy and Lin, Ming and Jiang, Chenfanfu and Gan, Chuang},
  journal={arXiv preprint arXiv:2303.05512},
  year={2023}
}

@inproceedings{wang2003multiscale,
  title={Multiscale structural similarity for image quality assessment},
  author={Wang, Zhou and Simoncelli, Eero P and Bovik, Alan C},
  booktitle={The thrity-seventh asilomar conference on signals, systems \& computers, 2003},
  volume={2},
  pages={1398--1402},
  year={2003},
  organization={Ieee}
}

@inproceedings{zhang2018unreasonable,
  title={The unreasonable effectiveness of deep features as a perceptual metric},
  author={Zhang, Richard and Isola, Phillip and Efros, Alexei A and Shechtman, Eli and Wang, Oliver},
  booktitle={Proceedings of the IEEE conference on computer vision and pattern recognition},
  pages={586--595},
  year={2018}
}

@article{cai2024gic,
  title={Gic: Gaussian-informed continuum for physical property identification and simulation},
  author={Cai, Junhao and Yang, Yuji and Yuan, Weihao and He, Yisheng and Dong, Zilong and Bo, Liefeng and Cheng, Hui and Chen, Qifeng},
  journal={Advances in Neural Information Processing Systems},
  volume={37},
  pages={75035--75063},
  year={2024}
}

@inproceedings{zhao2025toward,
  title={Toward Material-Agnostic System Identification from Videos},
  author={Zhao, Yizhou and Chen, Haoyu and Liu, Chunjiang and Li, Zhenyang and Herrmann, Charles and Hur, Junhwa and Li, Yinxiao and Yang, Ming-Hsuan and Raj, Bhiksha and Xu, Min},
  booktitle={Proceedings of the IEEE/CVF International Conference on Computer Vision},
  pages={5944--5956},
  year={2025}
}

\appendix

\end{document}